\documentclass[sigplan]{acmart}

\acmConference[EMC2 Workshop]{2019}{Arizona}

\setcopyright{none}

\usepackage{booktabs}   
\usepackage{comment}
\usepackage{soul}
\usepackage{url}
\usepackage{caption}
\usepackage{graphicx}
\usepackage{amsmath}
\usepackage{algorithm}
\usepackage{algorithmic}
\usepackage{multirow}
\begin{document}
\startPage{1}
\title{Run-Time Efficient RNN Compression for Inference on Edge Devices}         


\author{Urmish Thakker}
\affiliation{
  \position{Senior Research Engineer}
  \institution{Arm ML Research}            
}
\email{urmish.thakker@arm.com}          

\author{Jesse Beu}
\affiliation{
  \position{Staff Research Engineer}
  \institution{Arm ML Research}           
}
\email{jesse.beu@arm.com}         

\author{Dibakar Gope}
\affiliation{
  \position{Senior Research Engineer}
  \institution{Arm ML Research}           
}
\email{dibakar.gope@arm.com}         

\author{Ganesh Dasika}
\affiliation{
  \position{Principal Research Engineer}
  \institution{Arm ML Research}           
}
\email{ganesh.dasika@arm.com}     

\author{Matthew Mattina}
\affiliation{
  \position{Senior Director, ML and AI Research}
  \institution{Arm ML Research}           
}
\email{matthew.mattina@arm.com}  
\begin{abstract}
Recurrent neural networks can be large and compute-intensive, yet many applications that benefit from RNNs run on small devices with very limited compute and storage capabilities while still having run-time constraints. As a result, there is a need for compression techniques that can achieve significant compression without negatively impacting inference run-time and task accuracy. This paper explores a new compressed RNN cell implementation called Hybrid Matrix Decomposition (HMD) that achieves this dual objective. HMD creates dense matrices that results in output features where the upper sub-vector has "richer" features while the lower-sub vector has "constrained" features". On the benchmarks evaluated in this paper, this results in faster inference runtime than pruning and better accuracy than matrix factorization for compression factors of 2-4$\times$. 
\end{abstract}


\maketitle

\section{Introduction}

\label{intro}
Recurrent neural networks have shown state-of-the-art results for a wide variety of applications. Though many of these applications run on mobile devices, they are typically enabled by querying a cloud-based system to do most of the computation. The energy, latency, and privacy implications associated with running a query on the cloud is changing where users run a neural network application. We should, therefore, expect an increase in the number of RNNs running on embedded devices. Due to the energy and power constraints of edge devices, embedded SoCs frequently use lower-bandwidth memory technologies and smaller caches compared to desktop and server processors. Thus, there is a need for good compression techniques to enable large RNN models to fit into an edge device or ensure that they run efficiently on devices with smaller caches \cite{urmtha01RNN}. Additionally, compressing models should not negatively impact the inference run-time as these tasks may have realtime deadlines to provide a good user experience. 

In order to choose a compression scheme for a particular network, one needs to consider 3 different axes -- the compression factor, the speedup over the baseline, and the accuracy. Ideally, a good compression algorithm should not sacrifice improvement along one axis for improvement along another. For example, network pruning~\cite{han2015deep_compression} has shown to be an effective compression technique, but pruning creates a sparse matrix representation that is inefficient to execute on most modern CPUs. Our analysis shows that pruned networks can achieve a faster run-time than the baseline only for significantly high compression factors. Low-rank matrix factorization (LMF) is another popular compression technique that can achieve speedup proportional to the compression factor. However, LMF has had mixed results in maintaining model accuracy~\cite{lmf-good1,6638949,lmf-bad}. This is because LMF reduces the rank of a matrix significantly, reducing its expressibility. Lastly, structured matrices \cite{structuredmatrix,thakker2020compressing} can also be used to compress neural networks. While these techniques show a significant reduction in computation, this reduction only translates to a realized run-time improvement for larger matrices ~\cite{NIPS2018_8119} or while using specialized hardware~\cite{clstm-fpga}. 


\textbf{To overcome the problem of finding an alternative to pruning, when LMF leads to loss in accuracy, we introduce a new compression technique called Hybrid Matrix Decomposition (HMD) which can act as an effective compression technique for edge use cases.} The results are very promising -- HMD achieves iso-accuracy for a large compression factor (2$\times$ to 3$\times$), improves the run-time over pruning by a factor of 2$\times$, improves run-time over a structured matrix-based technique by a factor of $30\times$ and achieves better model accuracy than LMF.

The key contributions of this paper are:
\begin{itemize}
    \item Introduction of a new compression technique called Hybrid Matrix Decomposition that can regain most of the baseline accuracy at 2$\times$ to 3$\times$ compression factors.
    \item Comparison of the model accuracy, inference run-time and compression trade-offs of HMD with network pruning and matrix factorization
\end{itemize}

\section{Related Work}
The research in NN compression can be categorized under 4 topics - Pruning\cite{han2015deep_compression,suyog}, structured matrix based techniques \cite{circular2,structuredmatrix,Thakker2019PushingTL,urmishKron}, quantization \cite{gope2019ternary} and tensor decomposition \cite{tjandra2017compressing,DBLP:journals/corr/KuchaievG17}. HMD belongs in the structured matrix category. We compare our method against pruning, structured matrix and tensor decomposition techniques. Quantization is an orthogonal technique and can further compress the models presented in this paper.
\section{HMD-Based RNN Compression}
\label{sec:hybrid}

\begin{figure}[tb]
\vskip 0.1in
\centering
\includegraphics[width=0.7\columnwidth]{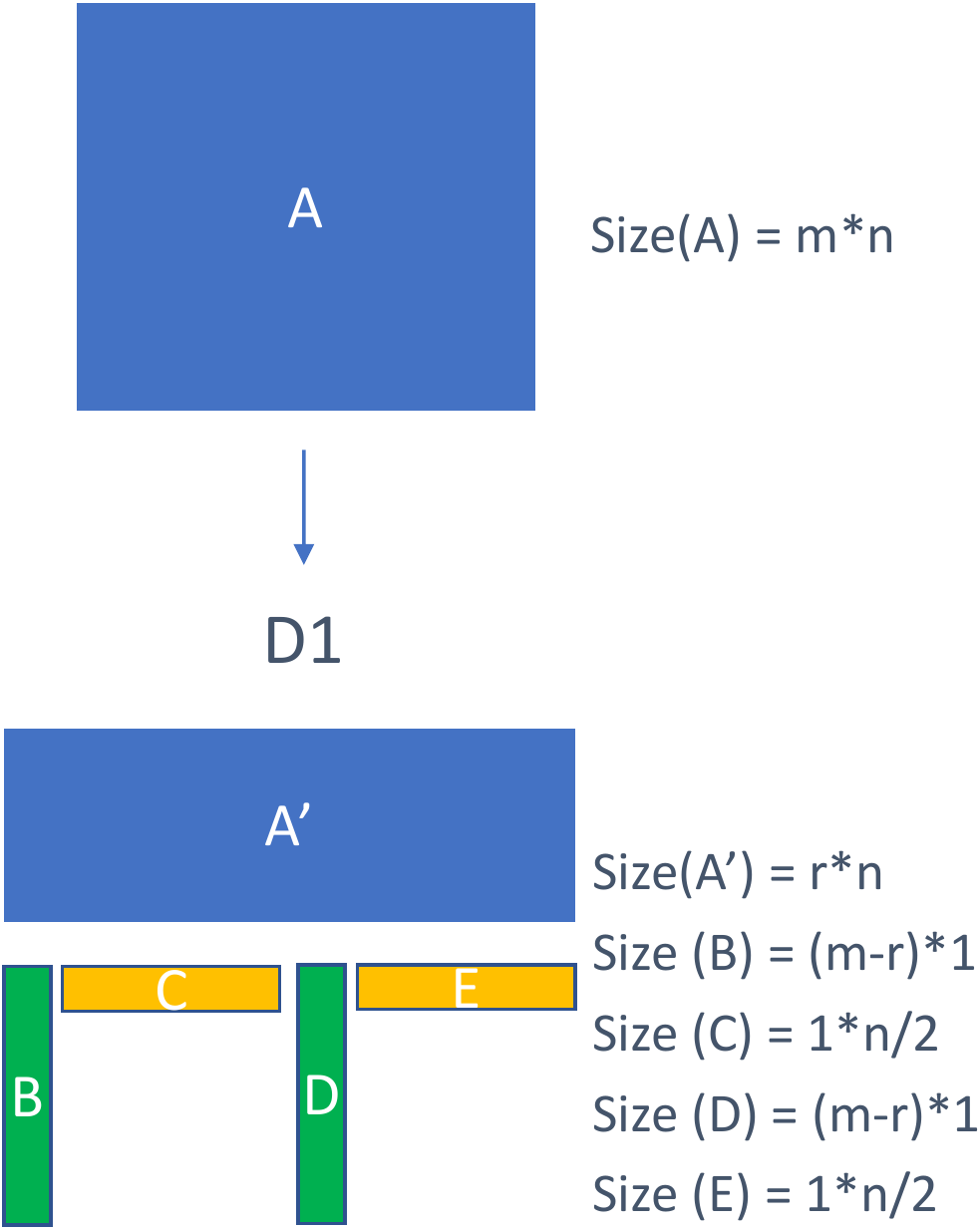}
\caption{Representation of a matrix using hybrid decomposition}
\label{hmatrix-fig}
\vskip -0.1in
\end{figure}

\begin{algorithm}[t]
   \caption{Reconstructing A in D1}
   \label{alg:hybrid-getA}
   \textbf{Input}: Matrices $A'$ of dimension $r \times n$, $B$ of dimension $(m-r) \times 1$, $C$ of dimension $1 \times (n/2)$, $D$ of dimension $(m-r) \times 1$, $E$ of dimension $1 \times (n/2)$ \\
   \textbf{Output}: Matrix $A$ of dimension $m\times n$ 
   \begin{algorithmic}[1]
   \STATE $G \gets B\times C$ 
   \STATE $H \gets D \times E$ 
   \STATE $K = concatenate(G,H,column)$ 
   \STATE $A =  concatenate(A',K,row)$
\end{algorithmic}
\end{algorithm}

The output of a RNN layer is a vector. Each element of the vector is derived from multiple fully connected layers followed by a non linearity operation. Thus, every element of an output vector is connected to every element of the input and hidden vectors of a RNN layer. This leads to a large number of parameters. Generally, not all elements of the output vector need to be connected this way to derive useful information from the input and the hidden vector. Pruning exploits these sparse connections in an unstructured manner. Additionally, most RNN networks are followed by a fully-connected softmax layer or another RNN layer. Even if the order of the elements in the output of a particular RNN layer changes, the weights in the subsequent fully connected or RNN layers can adjust to accommodate that. Thus, the order of the output vectors of RNN hidden layers is not strictly important.

These two properties of a RNN layer can be used to create a more hardware-friendly compression scheme. This paper introduces one such scheme -- Hybrid Matrix Decomposition. HMD splits the input and recurrent matrices in an RNN layer into two parts -- a fully parameterized upper part and  a lower part composed of rank-1 blocks. The upper part is used to generate elements of an output vector that need dense connectivity, while the lower part generates elements of the output vector that can generate useful information using sparse connectivity. There are multiple ways to constrain the lower part using rank-1 blocks. Figure \ref{hmatrix-fig} shows one such technique - D1. 


The D1 technique consists of an unconstrained upper half $A'$ and a constrained lower half. The lower half is composed of two rank-1 blocks. Algorithm \ref{alg:hybrid-getA} shows how to expand $A'$,$B$,$C$,$D$, and $E$ to get a matrix of size $m \times n$. In this paper, whenever we discuss HMD, we will refer to the D1 method to decompose the matrix. If we decompose the weight matrix using D1 technique, the storage reduction is given by:
\begin{equation}
\label{eq:hmd-d1}
\frac{m \times n}{(r \times n) + 2 \times (m - r + n/2) }
\end{equation}

\begin{algorithm}[t]
   \caption{Matrix vector product when a matrix uses the HMD technique as shown in D1}
   \label{alg:hybrid-mvalg}
   {\bfseries Input 1:} Matrices $A'$ of dimension $r \times n$, $B$ of dimension $(m-r) \times 1$, $C$ of dimension $1 \times (n/2)$, $D$ of dimension $(m-r) \times 1$, $E$ of dimension $1 \times (n/2)$ \\
   {\bfseries Input 2:} Vector $I$ of dimension $n \times 1$ \\
   {\bfseries Output:} Matrix $O$ of dimension $m\times1$ \\
   \begin{algorithmic}[1]
   \STATE $O\textsubscript{1:r} \gets A'\times I$
   \STATE $Temp1Scalar \gets C \times I\textsubscript{1:n/2} $
   \STATE $Temp1 \gets B \circ Temp1Scalar $
   \STATE $Temp2Scalar \gets E \times I\textsubscript{1+n/2:n} $
   \STATE $Temp2 \gets D \circ Temp2Scalar $
   \STATE $O\textsubscript{r+1:m} \gets Temp1 + Temp2$
   \STATE $O = concatenate\{O\textsubscript{1:r},O\textsubscript{r+1:m}\} $
\end{algorithmic}
\end{algorithm}

Apart from the storage reduction, HMD also leads to a reduction in the number of computations. Assuming a batch size of 1 during inference, Algorithm~\ref{alg:hybrid-mvalg} shows how to calculate the matrix vector product when the matrix is represented using HMD. This algorithm avoids expanding the matrix $A'$, $B$, $C$, $D$, and $E$ into $A$ as shown in Algorithm~\ref{alg:hybrid-getA} and uses the associative property of matrix products to gain the computation speedup. 
The compression in number of operations when we use Algorithm~\ref{alg:hybrid-mvalg} is:
\begin{equation}
\label{eq:hybrid-acr}
\frac{ m \times n  }{r \times n + 2 \times(n/2 + m-r) + m-r}  
\end{equation}

As discussed previously, HMD divides the output into two stacked sub-vectors: One is a result of a fully-parameterized multiplication ($A'\times I$) and the other is the result of the low rank multiplication ($C\times R \times I\textsubscript{1:n/2} + E \times F \times I\textsubscript{1+n/2:n}$). Thus, the upper sub vector has ``richer'' features while the lower sub vector has ``constrained'' features. 


\section{Results}
\label{sec:results}

We do an extensive comparison of HMD with 2 other compression techniques -- model pruning and matrix factorization. Additionally, we also compared HMD with a structured matrix-based compression technique called block circular decomposition (BCD)~\cite{clstm-fpga,circnn}. BCD-compressed networks were able to recover the baseline accuracy for $2\times$ - $4\times$ compression. However, the run-time of the compressed network was $30\times$ slower than baseline. As a result, we do not discuss the results using BCD compression in the rest of the paper.

Model pruning~\cite{suyog} induces sparsity in the matrices of a neural network
creating sparse matrices which are stored in a specialized CSR data structure.  The overhead of traversing these data structures while performing the matrix-vector multiplication can lead to poorer inference run-time than when executing the baseline, non-sparse network. 

Low Rank Matrix Factorization (LMF)~\cite{DBLP:journals/corr/KuchaievG17} expresses a larger matrix $A$ of dimension $m\times n$ as a product of two smaller matrices $U$ and $V$ of dimension $m\times d$ and $d\times n$, respectively. Parameter $d$ controls the compression factor.  Unlike pruning, Matrix Factorization is able to improve the run-time over the baseline for all compression factors. 
\subsection{Comparison of compression techniques across different ML tasks}
\label{sec:comparison}

The impact of compression on accuracy is compared for 3 benchmarks covering 2 different tasks -- Human Activity Recognition and Language Modeling. These tasks are some of the important applications that run on edge and embedded devices. In order to compare the inference run-time of RNN cells compressed using the 3 techniques discussed above, we implemented these cells in C++ using the Eigen library. 
We ran our experiments on a single cortex-A73 core  of the Hikey 960 board. The size of L3 cache is 2MB.

We compress the network using pruning, LMF, and HMD. Additionally, we train a smaller baseline with the number of parameters equal to that of the compressed baseline. 

\subsubsection{Human Activity Recognition (HAR)}
\label{sec:har}

We train two different networks for human activity recognition. Both of these networks are trained on the Opportunity dataset \cite{opp}. However, they differ in the way they process the dataset and the body sensors they chose to train their networks on. 

\textbf{HAR1}: The first HAR network is based on the work in \cite{hammerla2016deep}. The network uses a bidirectional LSTM with hidden length of size 179 followed by a softmax layer to get an accuracy of 91.9\%. Input is of dimension 77 and is fed over 81 time steps. The total number of parameters in this network are 374,468. 

\begin{figure}[t]
\vskip 0.1in
\begin{center}
\centerline{\includegraphics[width=\columnwidth]{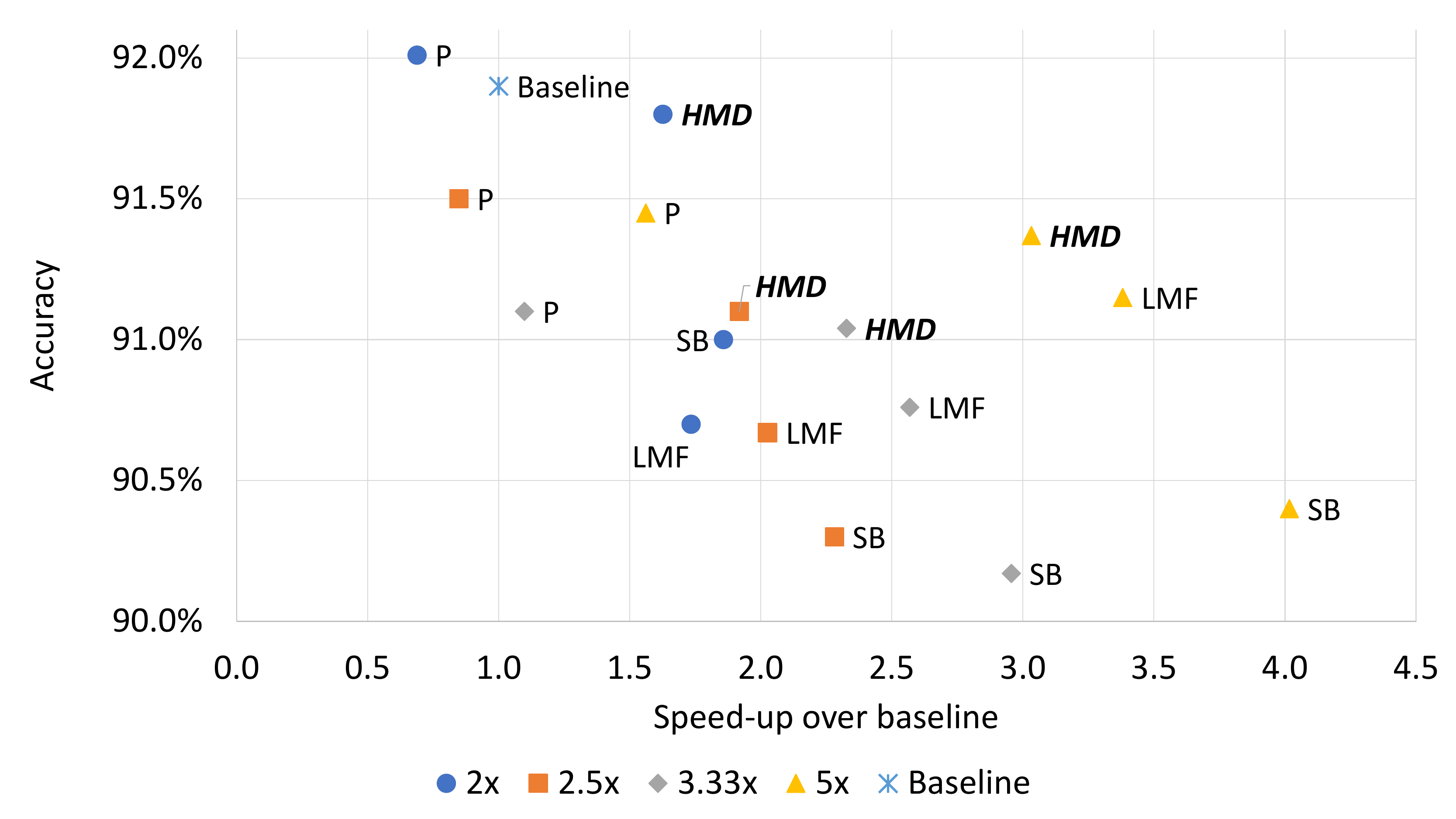}}
\caption{Accuracy vs speedup for the HAR1 network comparing the baseline with a smaller baseline and the baseline compressed using different compression schemes at varying compression factors. Speed-up values $>1$ indicate a decrease in inference run-time and values $<1$ indicate an increase in inference run-time. For each compression factor, the compression scheme that is most to the top-right is the ideal choice and is highlighted in bold italics. P = Pruning, LMF = Low rank matrix factorization, HMD = Hybrid matrix decomposition, SB = Smaller baseline.}
\label{fig:har1}
\end{center}
\vskip -0.1in
\end{figure}

Figure ~\ref{fig:har1} shows the result of compressing the LSTM layers in the baseline by $2\times$, $2.5\times$, $3.33\times$ and $5\times$. As we increase the compression, the accuracy degradation becomes larger for all compression schemes. Thus, the best compression scheme for each compression factor is a function of task accuracy and speedup required to run the application. For $2\times$ compression, HMD and pruning achieve better accuracy than the smaller baseline and LMF. Additionally, the HMD compressed network is $2\times$ faster than the pruned network. Similar observations can be made for $2.5\times$ and $3.33\times$ compression. Thus, HMD can be used as the preferred compression scheme for these compression factors. At $5\times$ compression, HMD is slightly more accurate than LMF while being 15\% slower. The preferred choice for compression scheme depends on what criteria (accuracy or speed) one is willing to sacrifice. Finally, all three compression schemes have better accuracy than the smaller baseline.

\textbf{HAR2}: The second HAR network is based on the work in \cite{deepConvLSTM}. They use 113 sensors from the Opportunity dataset. The network has 4 convolutional layers followed by 2 LSTM layers and a softmax layer. The total number of parameters in the network are 3,964,754. The LSTM layers are of size 128 contributing to more than 95\% of the total parameters. 

\begin{figure}[t]
\vskip 0.1in
\begin{center}
\centerline{\includegraphics[width=\columnwidth]{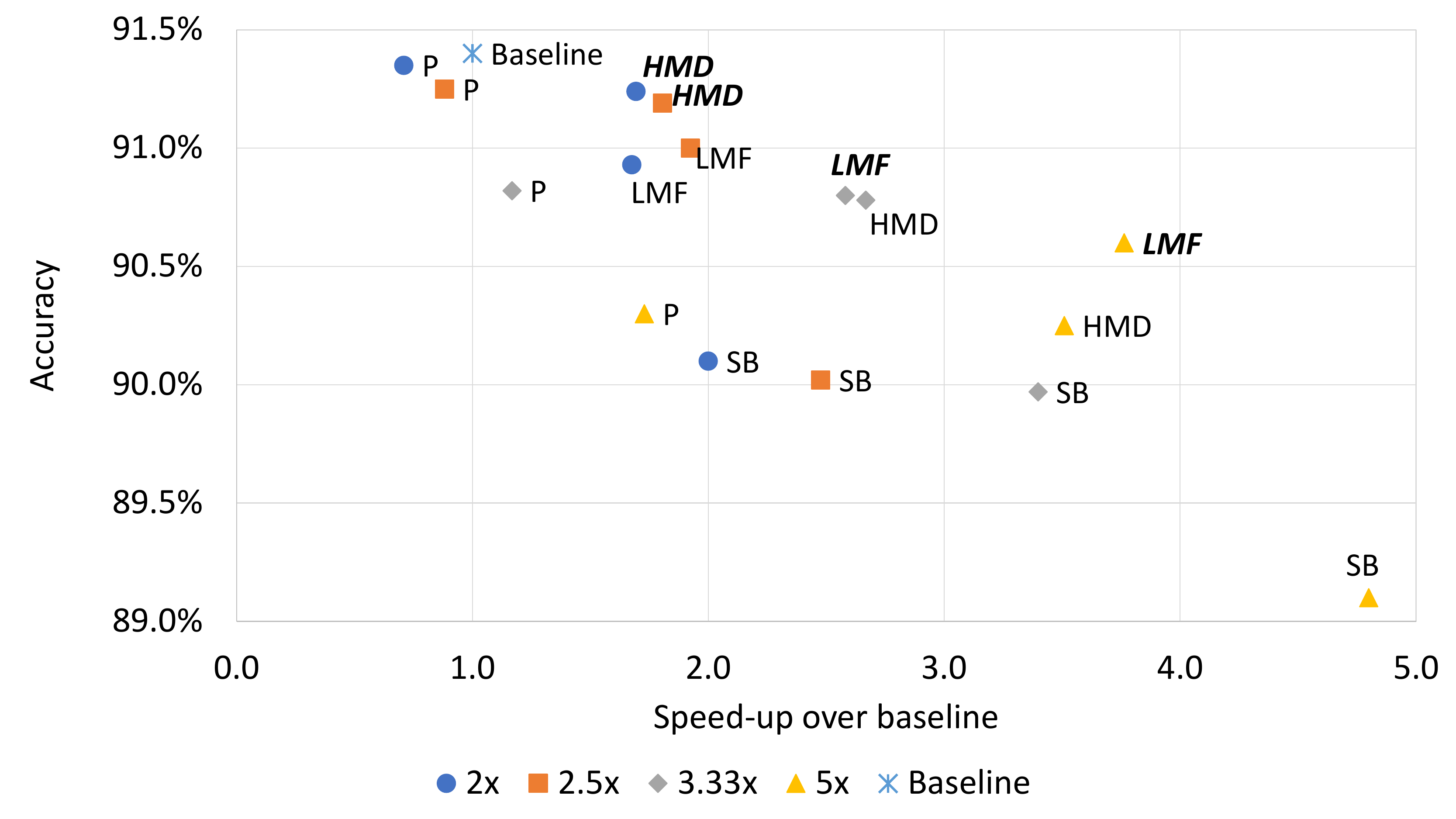}}
\caption{Accuracy vs speedup for HAR2 network comparing the baseline with a smaller baseline and the baseline compressed using different compression schemes at varying compression factors. Speed-up values $>1$ indicate a decrease in inference run-time and values $<1$ indicate an increase in inference run-time. For each compression factor, the compression scheme that is most to the top-right is the ideal choice and is highlighted in bold italics. P = Pruning, LMF = Low rank matrix factorization, HMD = Hybrid matrix decomposition, SB = Smaller baseline.}
\label{fig:har2}
\end{center}
\vskip -0.2in
\end{figure}

Figure ~\ref{fig:har2} shows the result of compressing the LSTM layers in baseline by $2\times$, $2.5\times$, $3.33\times$ and $5\times$. As we increase the compression, the accuracy degradation becomes larger for all compression schemes. For $2\times$ and $2.5\times$ compression factors, HMD is the superior technique, achieving better run-time than pruning ($2\times$ faster) and better accuracy than LMF (improvement of $0.4\%$). 
For higher compression factors, LMF becomes an attractive option to compress the HAR2 application. For $3.33\times$ compression, LMF, HMD and pruning achieve equivalent accuracy. However, LMF is slightly faster than HMD and more than $2\times$ faster than pruning. Finally, all three compression schemes have better accuracy than the smaller baseline.

\subsubsection{Language Modeling}
\label{sec:ptblm}
We use the small model from ~\cite{ptblm} as our baseline. The baseline has 2 LSTM layers each with a hidden vector of size 200. Additionally, it uses 10,000 words from the English vocabulary. Together with the input and output word embeddings, the total size of the network is 4,171,000 parameters.

\begin{figure}[t]
\vskip 0.1in
\begin{center}
\centerline{\includegraphics[width=\columnwidth]{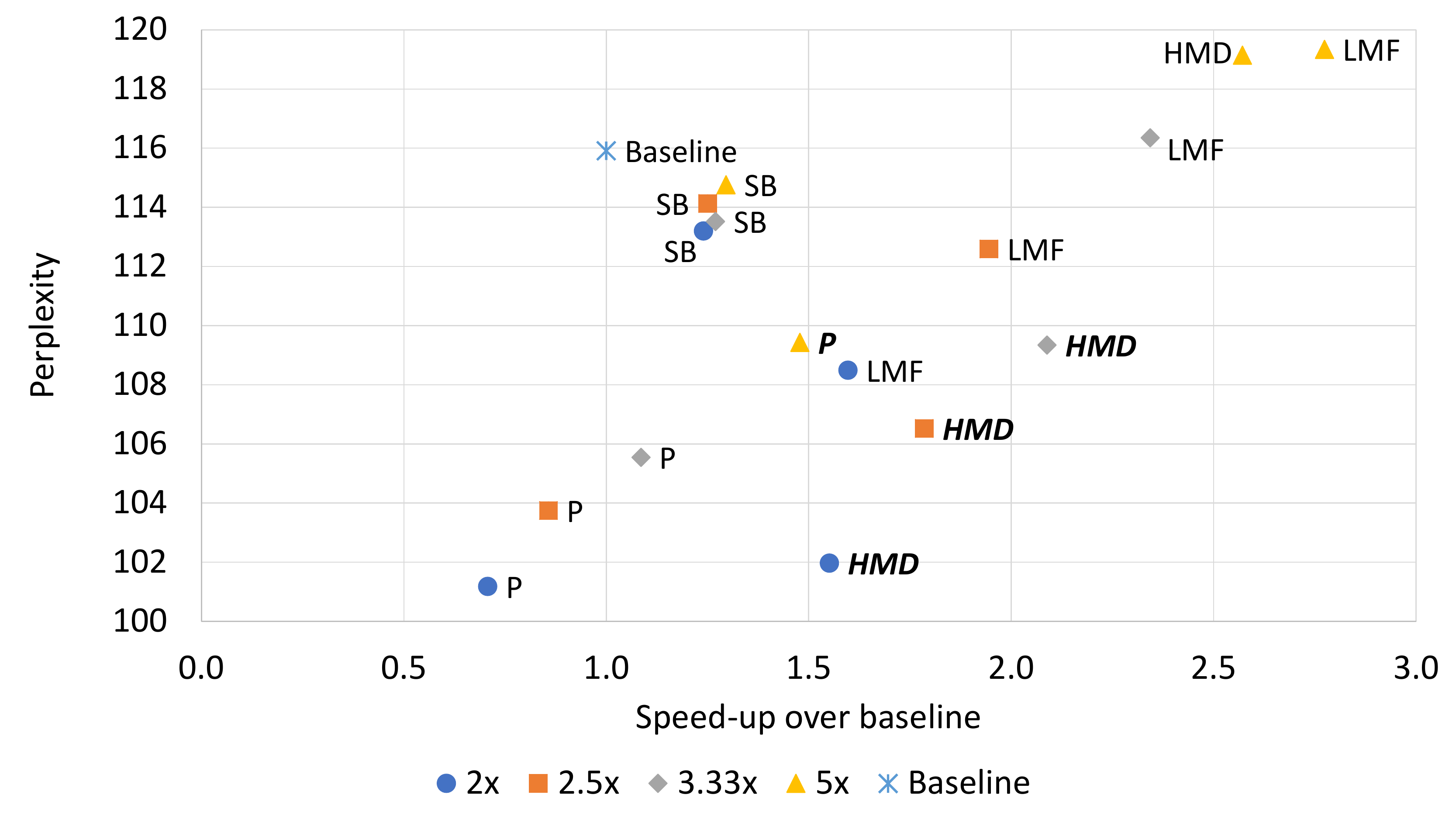}}
\caption{Perplexity vs speedup for PTB-LM network comparing the baseline with a smaller baseline and the baseline compressed using different compression schemes at varying compression factors. Speed-up values $>1$ indicate a decrease in inference run-time and values $<1$ indicate an increase in inference run-time. In case of perplexity, lower values are better. Thus, for each compression factor, the compression scheme that is most to the bottom-right is the ideal choice and is highlighted in bold italics. P = Pruning, LMF = Low rank matrix factorization, HMD = Hybrid matrix decomposition, SB = Smaller baseline.}
\label{fig:ptblm}
\end{center}
\vskip -0.1in
\end{figure}

Figure ~\ref{fig:ptblm} shows the results of compressing the LSTM layers in the baseline by $2\times$, $2.5\times$, $3.33\times$ and $5\times$.  In case of LM, lower the perplexity, better the model. Pruning consistently achieves better accuracy than baseline and other compression techniques. However, pruning never achieves a better speedup than other compression techniques 
LMF achieves better perplexity than baseline for $2\times$ and $2.5\times$ compression and achieves speedup for all compression factors. But LMF, does not beat the perplexity values achieved by HMD. HMD simultaneously achieves better perplexity than baseline for most compression factor, better perplexity than LMF for all compression factors and faster inference run-time than baseline and pruned networks for all compression factors. 
Thus, HMD makes a strong case for being the preferred compression scheme

\section{Conclusion}
Choosing the right compression technique requires looking at three criteria -- compression factor, accuracy, and run-time. Pruning is an effective compression technique, but can sacrifice speedup over baseline for certain compression factors. LMF achieves better speedup than baseline for all compression factors, but can lead to accuracy degradation. This paper introduces a new compression scheme called HMD, which is extremely effective when compression using pruning does not lead to speedup over baseline and LMF leads to accuracy degradation. 

\bibliography{HMD}

\end{document}